\documentclass{article}
\usepackage{spconf,amsmath,graphicx}
\usepackage{enumitem}
\usepackage{algorithm}
\usepackage{algorithmic}
\usepackage{amsfonts}  
\usepackage{multirow}
\usepackage{array}
\usepackage{caption}
\usepackage{pgfplots}
\pgfplotsset{compat=1.17}
\usepackage{arydshln}
\usepackage{subcaption}
\usepackage[numbers]{natbib}
\title{Frequency Masking for Universal Deepfake Detection}
\name{Chandler Timm Doloriel \qquad Ngai-Man Cheung\sthanks{Corresponding Author}} 
\address{Singapore University of Technology and Design (SUTD)}
\begin{document}
%
\maketitle
\begin{abstract}
We study universal deepfake detection. Our  goal is to detect synthetic images from a range of generative AI approaches, particularly from emerging ones which are unseen during training of the deepfake detector. Universal deepfake detection requires outstanding generalization capability. Motivated by recently proposed masked image modeling which has demonstrated excellent generalization in self-supervised pre-training, we make the first attempt to explore masked image modeling for universal deepfake detection. We study spatial and frequency domain masking in training deepfake detectors. Based on empirical analysis, we propose a novel deepfake detector via frequency masking. Our focus on frequency domain is different from the majority, which primarily target spatial domain detection. Our comparative analyses reveal substantial performance gains over existing methods. Code and models are publicly available\footnote{https://github.com/chandlerbing65nm/FakeImageDetection.git}.
\end{abstract}
\vspace{-0.3cm}
\section{Introduction}
\label{sec:intro}

The proliferation of increasingly convincing synthetic images, facilitated by generative AI, poses significant challenges across multiple sectors, including cybersecurity, digital forensics, and public discourse~\cite{ojha2023fakedetect, Gragnaniello2021, DBLP:conf/eccv/ChaiBLI20, Abdollahzadeh2023ASO}. 
These AI-generated images could be mis-used as deepfakes for malicious purpose, e.g., disinformation. Detection of deepfakes is an important problem that  has attracted attention.

{\bf Universal deepfake detection.}
Early efforts in deepfake detection focus on identifying synthetic images generated by particular types of generative AI~\cite{mirsky2020deepfake}. However, with rapid advancements (e.g. diffusion models~\cite{rombach2021diffusion}), there is an increasing interest in studying {\em  universal deepfake detection}  capable of performing effectively for a range of generative AI approaches,  particularly for emerging ones which  are {\em unseen} during training of the deepfake detector. Therefore, universal deepfake detection necessitates a significant generalization capability. Wang et al.~\cite{wang2019cnngenerated} investigated post-processing and data augmentation techniques for detecting synthetic images, primarily those generated by various Generative Adversarial Network (GAN) models. Subsequently, Chandrasegaran et al.~\cite{Chandrasegaran_2022_ECCV} focused on forensic features for universal deepfake detection. More recent work, as demonstrated by Ojha et al.~\cite{ojha2023fakedetect}, involve utilizing the feature space of a large pretrained model for this purpose. However, this method hinges on the availability of a model pretrained on a very large dataset. Finally, Chen et al.~\cite{DBLP:conf/nips/ChenZSWL22} investigated the application of one-shot test-time training to enhance generalization in detection, albeit at the cost of additional computational resources for each test sample.



{\bf Masked image modeling.}
Meanwhile, in self-supervised pre-training, masked image modeling has emerged in the past year as a  promising approach to improve generalization capability~\cite{DBLP:conf/cvpr/HeCXLDG22,DBLP:journals/corr/abs-2302-02615,
DBLP:conf/iclr/0002LZ0OL23,DBLP:conf/nips/HuangCGXZLLX22}.
Specifically, in pre-training, the pre-text tasks learn to predict masked portions of the unlabeled data, and reconstruction loss is used as the learning objective. After pre-training, the pre-trained encoder can be effectively adapted for various downstream tasks. The study presented in~\cite{DBLP:conf/cvpr/HeCXLDG22} demonstrates that pre-training through a masked autoencoder (MAE) empowers high-capacity models to attain state-of-the-art (SOTA) generalization performance across a multitude of downstream tasks. Furthermore, extensive experiments outlined in~\cite{DBLP:journals/corr/abs-2302-02615} reveal that employing masking techniques can lead to SOTA results in out-of-distribution detection.

\begin{figure}[t!]

    \begin{subfigure}[b]{\linewidth}
        \centering
        \includegraphics[width=0.90\linewidth]{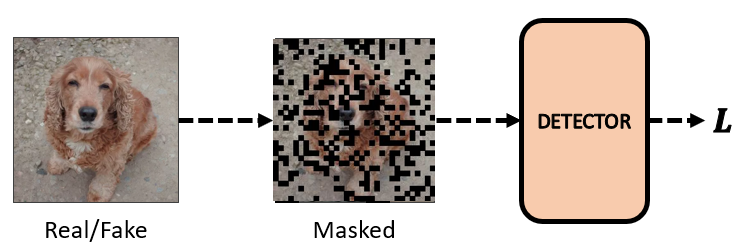}
        \caption{Spatial Domain Masking}
        \label{fig:sub1}
    \end{subfigure}
    
    \vspace{1em} 
    
    \begin{subfigure}[b]{\linewidth}
        \centering
        \includegraphics[width=0.90\linewidth]{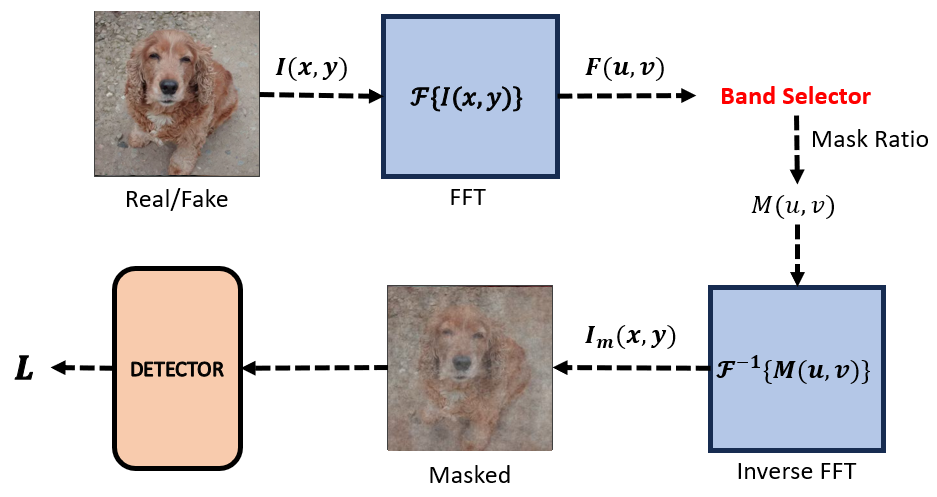}
        \caption{Frequency Domain Masking}
        \vspace{5pt}
        \label{fig:sub3}
    \end{subfigure}
    
    \caption{Our proposed training of universal deepfake detector using  spatial and frequency domain masking.
    In both cases, \( L \) represents the binary cross-entropy loss for real/fake discrimination. (a) Our spatial domain masking uses either individual pixels or patches to mask portions of the input image. (b) Our frequency domain masking  transforms the input image \( I(x, y) \) to the frequency domain \( F(u, v) \) using FFT. Guided by a frequency band selector and mask ratio, specific frequencies within \( F(u, v) \) are nullified to yield \( M(u, v) \). The inverse FFT produces the masked image \( I_{\text{m}}(x, y) \), serving as the classifier input for training universal deepfake detection. We remark that masking is  applied only in the supervised training stage to encourage the detector to learn generalizable representation. No masking is applied in the testing stage.}
    \label{fig:global}
\end{figure}

{\bf In our work,} we make the first attempt to explore masked image modeling to improve generalization capability of deepfake detector with the objective to advance universal deepfake detection. Unlike traditional masked image modeling which primarily uses reconstruction loss in self-supervised pre-training, our method applies masking in a supervised setting, focusing on classification loss for distinguishing real and fake images. Our training involves both spatial and frequency domain masking on all images, as depicted in Figure~\ref{fig:global}. This technique, which obscures parts of the image, enhances the challenge of training. It aims to prevent the detector from depending on superficial features and instead fosters the development of robust, generalizable representations for effective universal deepfake detection. Notably, masking is only employed during training, not in the testing phase.





In our study, we analyze both spatial and frequency masking for universal deepfake detection.
Our results suggest that frequency masking is more effective than spatial masking in generalizing deepfake detection for different generative AI approaches.
Our finding is consistent with a recent study by 
Corvi et al.~\cite{Corvi_2023_CVPR}, which discovers frequency artifacts in GAN and diffusion model-based synthetic images.
Different from~\cite{Corvi_2023_CVPR}, our main contribution is a new training method to improve detection accuracy via frequency masking.  
We remark that most 
existing detectors focus on spatial domain~\cite{ojha2023fakedetect, Gragnaniello2021, wang2019cnngenerated}. Our contributions are summarized as follows:

\begin{enumerate}[noitemsep, topsep=1pt, parsep=1pt, partopsep=1pt]
    \item We present the first study to explore masked image modeling for universal deepfake detection.
    \item We analyze 
    two distinct types of masking methods (spatial,  frequency), and empirically demonstrate that frequency masking performs better (Fig.~\ref{fig:global}). 
    \item We conduct analysis and experiments to validate the effectiveness of universal deepfake detection via frequency masking.
\end{enumerate}

\vspace{-0.4cm}
\section{Methodology}

To improve the performance of universal deepfake detection systems, we introduce masking strategies operating in various domains—spatial and frequency. This section delves into details of these techniques, showing how they contribute to improvement of universal deepfake detection.


\vspace{-0.4cm}
\subsection{Spatial Domain Masking}
\label{subsec:math_spatial_patch_masking}

For spatial domain masking, we study two distinct methods: Patch Masking and Pixel Masking. 
\textbf{Patch Masking}  operates by dividing an image of size \(H \times W\) into non-overlapping patches of size \(p \times p\). The number of patches \(N\) can be calculated:
$N = \left\lfloor {H \times W}/{p^2} \right\rfloor$.
A ratio \(r\) is used to determine the number of patches \(m\) that will be masked, calculated as \(m = \lceil r \times N \rceil\). These masked patches are selected randomly, and their pixel values are set to zero.

In \textbf{Pixel Masking}, each pixel is independently considered. Given an image of dimensions \(H \times W\) , the total number of pixels is \(T = H \times W\). A ratio \(r\) specifies the portion of pixels to mask, resulting in \(m = \lceil r \times T \rceil\) masked pixels. These are selected randomly across the image.

The masking operation for both patch and pixel masking methods can be defined as:

\begin{equation}
M(i, j) = \begin{cases}
0 & \text{if }(i, j) \in m \\
I(i, j) & \text{otherwise}
\end{cases}
\end{equation}
where \(M\) is the masked image and \(I\) is the original image.
Both methods create a binary mask that is element-wise multiplied with the original image to produce the masked image. These masking strategies serve as the foundation for our frequency-based masking approach, enabling effective universal deepfake detection by focusing on important features.




\vspace{-0.4cm}
\subsection{Frequency Domain Masking}
\label{subsec:math_frequency_masking}

Our frequency domain masking utilizes Fast Fourier Transform (FFT) to represent the image in terms of its frequency components. Given an image of dimensions \(H \times W \) where \(H\) and \(W\) are the height and width respectively, we first compute its frequency representation \(F(u, v)\) using the FFT:

\begin{equation}
F(u, v) = \mathcal{F}\{ I(x, y) \}
\end{equation}

Here \(u\) and \(v\) corresponds to the frequency along the image's width and height, respectively. \( \mathcal{F} \) denotes the FFT operation, and \( I(x, y) \) is the original image in spatial coordinates.

The frequency masking is dictated by a masking ratio \( r \) and a specified frequency band (`low', `mid', `high', `all'). The regions for each frequency band are defined as follows:

\begin{itemize}
  \item \textbf{Low Band:} \(0 \leq u < \frac{H}{4}, 0 \leq v < \frac{W}{4}\)
  \item \textbf{Mid Band:} \(\frac{H}{4} \leq u < \frac{3H}{4}, \frac{W}{4} \leq v < \frac{3W}{4}\)
  \item \textbf{High Band:} \(\frac{3H}{4} \leq u < H, \frac{3W}{4} \leq v < W\)
  \item \textbf{All Band:} \(0 \leq u < H, 0 \leq v < W\)
\end{itemize}

The division of frequency bands into \textbf{Low}, \textbf{Mid}, and \textbf{High} serves to isolate the contributions of specific frequency components to the overall image features. These divisions are calculated based on the dimensions \(H \times W\) of the Fourier transform of the image. The \textbf{Low Band}, captures the coarse or global features of an image. These are the low-frequency components that represent the most significant portions of the image. The \textbf{Mid Band} targets medium-frequency components, which often account for textures and other finer details. Finally, the \textbf{High Band} focuses on high-frequency noise and edge details, which are less dominant but possibly crucial for tasks like deepfake detection. The \textbf{All Band} simply includes all the frequencies, providing the most comprehensive masking strategy.

Given the selected frequency band \([u_{\text{start}}, u_{\text{end}}] \times [v_{\text{start}}, v_{\text{end}}]\), the frequencies to be nullified \(N\) is calculated:
$N = \lceil r \times (u_{\text{end}} - u_{\text{start}}) \times (v_{\text{end}} - v_{\text{start}}) \rceil$.
Frequencies within this region are set to zero, resulting in a masked frequency representation \(M(u, v)\):

\begin{equation}
M(u, v) = \begin{cases} 
0 & \text{if } (u, v)  \in N \\
F(u, v) & \text{otherwise}
\end{cases}
\end{equation}
The masked image \(I_{\text{m}}(x, y)\) is obtained by applying the inverse FFT to \(M(u, v)\):
$I_{\text{m}}(x, y) = \mathcal{F}^{-1}\{ M(u, v) \}$.


\begin{figure}[t]
    \begin{center}
        \begin{tikzpicture}[scale=0.6]
            \begin{axis}[
                ybar,
                xlabel={Mask Types},
                ylabel={Mean Average Precision (\%)},
                xmin=0.8, xmax=3.2,
                ymin=74, ymax=90,
                xtick={1,2,3},
                xticklabels={Pixel, Patch, Frequency},
                ymajorgrids=true,
                grid style=dashed,
                bar width=0.8cm, 
                enlarge x limits={abs=0.3}, 
                legend pos=north west,
            ]
            
            \addplot[
                fill=blue!40,
                ] 
                coordinates {
                (1,75.12)(2,86.09)(3,88.22)
                };
                \legend{mAP}
            \end{axis}
        \end{tikzpicture}
    \end{center}
    \caption{Performance of different masking types in terms of mean Average Precision (mAP) at 15\% masking ratio. The graph shows a marked improvement when transitioning from Pixel to Patch, and eventually to Frequency-based masking. Specifically, Frequency-based masking attains the highest mAP of 88.22\%, underscoring its effectiveness over the other masking types.}
    \label{fig:masking_types}
\end{figure}
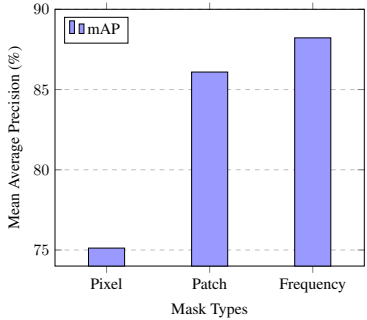

\begin{table}[ht!]
    \centering
    \label{tab:map_bargraph}
    \resizebox{0.95\linewidth}{!}{\tiny
    \begin{tabular}{c|c}
        \hline
        Masking Ratio (\%) & Mean Average Precision (\%) \\
        \hline
        0\%  & 85.86 \\
        15\% & \textbf{88.22} \\
        30\% & 87.20 \\
        50\% & 85.12 \\
        70\% & 83.86 \\
        \hline
    \end{tabular}}
    \caption{The table presents mean average precision (mAP) scores for varying degrees of frequency masking. The highest mAP is achieved at a 15\% masking ratio.}
\end{table}

\begin{table}[t!]
\label{table:freqbands}
\centering
\resizebox{0.98\linewidth}{!}{\tiny
\begin{tabular}{ccccc}
\hline
\multicolumn{1}{c}{\multirow{2}{*}{\parbox{1.5cm}{\centering Generative Models}}} & \multicolumn{4}{c}{Masked Frequency Bands} \\
\cline{2-5}
\multicolumn{1}{c}{} & Low & Mid & High & All \\
\hline
 GANs & 95.73 & 93.92 & 94.94 & \textbf{96.16} \\
 DeepFake & 85.64 & \textbf{87.22} & 80.17 & 79.07 \\
 Low-Level Vision & 85.77 & 83.69 & \textbf{88.78} & 87.27 \\
 Perceptual Loss & \textbf{99.21} & 97.98 & 98.11 & 98.41 \\
 Guided Diffusion & \textbf{74.90} & 70.19 & 69.26 & 72.42 \\
 Latent Diffusion & 76.35 & 75.60 & 66.39 & \textbf{80.45} \\
 Glide & \textbf{85.69} & 80.65 & 73.66 & 84.98 \\
 DALL-E & 70.60 & 68.72 & 71.18 & \textbf{80.11 }\\
\hline
 average mAP & 87.45 & 85.35 & 83.38 & \textbf{88.22} \\
\hline
\end{tabular}}
\caption{Scores of mean Average Precision (mAP) across various generative models for random 15\% masking of different frequency bands.}
\end{table}

\begin{table*}[t]
\label{table:compararch}
\renewcommand{\arraystretch}{1.7}  
\centering
\resizebox{\linewidth}{!}{\huge
\begin{tabular}{cc|ccccccccccccccccccc|c}
\hline
\multirow{2}{*}{\centering Method} & \multirow{2}{*}{Variant} & \multicolumn{6}{c}{Generative Adversarial Networks} & \multirow{2}{*}{\parbox{1.5cm}{\centering  Deep\\Fake}} & \multicolumn{2}{c}{Low-Level Vision} & \multicolumn{2}{|c}{Perceptual Loss} & \multirow{2}{*}{\parbox{2.2cm}{\centering Guided}} & \multicolumn{3}{c|}{LDM} & \multicolumn{3}{c}{Glide} & \multirow{2}{*}{\parbox{2.8cm}{\centering DALL-E}} & average \\
\cline{3-8} \cline{10-13} \cline{15-20}
  &  & \parbox{1.5cm}{\centering Pro\\GAN} & \parbox{1.5cm}{\centering Cycle\\GAN} & \parbox{1.5cm}{\centering Big\\GAN} & \parbox{1.5cm}{\centering Style\\GAN} & \parbox{1.5cm}{\centering Gau\\GAN} & \parbox{1.5cm}{\centering Star\\GAN} &  & \parbox{1.5cm}{\centering SITD} & \parbox{1.5cm}{\centering SAN} & \parbox{1.5cm}{\centering CRN} & {\centering IMLE} &  & \parbox{1.5cm}{\centering 200\\steps} & \parbox{2.4cm}{\centering 200\\w/ CFG} & \parbox{1.5cm}{\centering 100\\steps} & \parbox{1.5cm}{\centering 100\\27} & \parbox{1.5cm}{\centering 50\\27} & \parbox{1.5cm}{\centering 100\\10} &  & mAP \\
\hline
  Wang et al. ~\cite{wang2019cnngenerated} & Blur+JPEG (0.5) & 100.00 & 96.83 & 88.24 & 98.51 & 98.09 & 95.45 & 66.27 & 92.76 & 63.87 & 98.94 & 99.52 & 68.35 & 65.92 & 66.74 & 65.99 & 72.03 & 76.52 & 73.22 & 66.26 & 81.76 \\
  Wang et al. ~\cite{wang2019cnngenerated} + Ours & Blur+JPEG (0.5) & 100.00 & 92.97 & 90.06 & 98.15 & 97.80 & 87.94 & 73.09 & 89.92 & 74.57 & 96.97 & 98.00 & 70.98 & 77.75 & 72.44 & 78.21 & 78.53 & 82.61 & 79.27 & 78.16 & \textbf{85.13} (\textcolor{red}{\textbf{+3.37}}) \\
\hdashline
  Wang et al.~\cite{wang2019cnngenerated} & Blur+JPEG (0.1) & 100.00 & 93.47 & 84.51 & 99.59 & 89.49 & 98.15 & 89.02 & 97.23 & 70.45 & 98.22 & 98.39 & 77.67 & 71.16 & 73.01 & 72.53 & 80.51 & 84.62 & 82.06 & 71.30 & 85.86 \\
  Wang et al.~\cite{wang2019cnngenerated} + Ours & Blur+JPEG (0.1) & 100.00 & 93.75 & 92.19 & 98.93 & 97.18 & 94.93 & 79.07 & 95.72 & 78.81 & 98.63 & 98.19 & 72.42 & 81.89 & 78.01 & 81.45 & 83.13 & 87.50 & 84.30 & 80.11 & \textbf{88.22} (\textcolor{red}{\textbf{+2.36}}) \\
\hdashline
  Gragnaniello et al.~\cite{Gragnaniello2021} & Blur+JPEG (0.1) & 100.00 & 83.24 & 94.29 & 99.95 & 91.69 & 99.99 & 91.23 & 92.75 & 73.67 & 98.19 & 97.85 & 78.31 & 88.37 & 88.15 & 89.26 & 89.41 & 93.39 & 90.48 & 92.46 & 91.19 \\
  Gragnaniello et al.~\cite{Gragnaniello2021} + Ours & Blur+JPEG (0.1) & 100.00 & 94.07 & 97.61 & 99.92 & 98.52 & 99.99 & 94.01 & 95.44 & 81.64 & 96.73 & 95.47 & 80.10 & 94.60 & 94.46 & 95.02 & 90.68 & 93.72 & 91.42 & 96.47 & \textbf{94.20} (\textcolor{red}{\textbf{+3.01}}) \\
\hdashline
  Ojha et al.~\cite{ojha2023fakedetect} & Blur+JPEG (0.1) & 100.00 & 99.16 & 96.08 & 90.72 & 99.81 & 98.67 & 77.30 & 67.21 & 74.81 & 72.97 & 94.02 & 81.42 & 95.45 & 80.68 & 96.44 & 90.53 & 91.54 & 90.09 & 86.59 & 88.60 \\
  Ojha et al.~\cite{ojha2023fakedetect} + Ours & Blur+JPEG (0.1) & 100.00 & 99.04 & 96.89 & 92.66 & 99.83 & 98.86 & 77.17 & 67.41 & 75.80 & 78.83 & 95.85 & 81.77 & 95.45 & 80.93 & 96.48 & 90.66 & 91.71 & 90.23 & 86.73 & \textbf{89.28} (\textcolor{red}{\textbf{+0.68}}) \\
\hline
\end{tabular}}
\caption{Generalization results showcasing the effectiveness of our frequency-based masking technique. We compare the Average Precision (AP) with Wang et al.~\cite{wang2019cnngenerated}, Gragnaniello et al.~\cite{Gragnaniello2021}, and Ojha et al.~\cite{ojha2023fakedetect}, which are SOTA according to a recent study~\cite{Corvi_2023_ICASSP}. When integrated with these SOTA methods, our approach consistently improves mAP by significant amounts, highlighting the utility of frequency masking in enhancing universal deepfake detection.}
\end{table*}

\vspace{-0.4cm}
\section{Experiments}
\label{sec:typestyle}
\vspace{-0.2cm}


\textbf{Dataset}: In our experiments, we followed strictly the training and validation setup from from Wang et al.~\cite{wang2019cnngenerated} using ProGAN with 720k and 4k samples, respectively. For testing, we utilized data from models such as GANs (ProGAN, CycleGAN, BigGAN, StyleGAN, GauGAN, and StarGAN), DeepFake, low-level vision models (SITD and SAN), and perceptual loss models (CRN and IMLE) from Wang et al.~\cite{wang2019cnngenerated}. Additionally, we incorporated 1k samples per class from diffusion models~\cite{ojha2023fakedetect}: Guided Diffusion, Latent Diffusion (LDM) with varying steps of noise refinement (e.g., 100, 200) and generation guidance (w/ CFG), Glide which used two stages of noise refinement steps. The first stage (e.g., 50, 100) is to get low-resolution \(64 \times 64\) images and then use 27 steps to upsample the image to \(256 \times 256\) in the next stage, and lastly DALL-E-mini.


\textbf{State-of-the-art (SOTA)}: 
We compared with following SOTA: 
Wang et al.~\cite{wang2019cnngenerated} and Gragnaniello et al.~\cite{Gragnaniello2021}. These methods achieve SOTA detection accuracy across GANs and diffusion models according to a recent study~\cite{Corvi_2023_ICASSP}. We applied our proposed frequency masking on these SOTA and evaluated the improvement. 

As our default setting, we used Wang et al.~\cite{wang2019cnngenerated} and incorporated image augmentations like Gaussian Blur and JPEG compression, each having a 10\% likelihood of being applied.




\vspace{-0.4cm}
\subsection{Comparison of Mask Types}

For masking types, as depicted in Figure~\ref{fig:masking_types}, we observe a distinct hierarchy in terms of mean average precision (mAP). Pixel masking exhibits the lowest performance with an mAP of 75.12\%. In contrast, patch masking
with random 8x8 blocks of pixels as patches 
shows a notable improvement with an mAP of 86.09\%. However, the most interesting result is the superior performance of frequency-based masking, which reaches an mAP of 88.22\%. This clearly indicates that frequency-based masking captures more generalizable features compared to pixel and patch masking, thereby substantiating its effectiveness in universal deepfake detection task.

\vspace{-0.5cm}
\subsection{Ratio of Frequency Msking}

As shown in Table 1, we observe a clear trend in the performance of our frequency masking technique across different masking ratios. The highest mean average precision (mAP) of \(88.22\%\) is achieved at a masking ratio of \(15\%\). As the masking ratio increases, there is a noticeable decline in performance, with the mAP dropping to \(83.86\%\) at a \(70\%\) masking ratio. This trend suggests that the model is sensitive to the proportion of frequencies being masked, and excessive masking can compromise the model's ability to detect subtle features in the images. Therefore, based on these results, we select \(15\%\) as the default masking ratio for frequency masking in our experiments.

\vspace{-0.4cm}
\subsection{Frequency Bands for Masking}

Table 2 presents the mAP scores for various generative models when different frequency bands—Low, Mid, High, and All—are randomly masked. The highest mAP is observed when all frequency bands are masked (\textbf{88.22\%}), suggesting that a comprehensive masking strategy might be most effective for universal deepfake detection. However, there are nuances; for instance, DeepFake performs best with Mid-frequency masking (\textbf{87.22\%}), while Low-Level Vision models excel at High-frequency masking (\textbf{88.78\%}). This variance indicates that different generative models may leave distinct forensic artifacts in different frequency bands, making a targeted masking strategy potentially beneficial.

\vspace{-0.4cm}
\subsection{Comparison with State-Of-The-Art}



As shown in Table 3, `+ Ours' approach, incorporating \(15\%\) random masking across `all' frequency bands, consistently enhances performance when integrated with existing state-of-the-art (SOTA) methods. Specifically, the combination of our frequency-based masking technique with Wang et al.'s method~\cite{wang2019cnngenerated} results in an increase of \(3.37\%\) and \(2.36\%\) in mean average precision (mAP) for the Blur+JPEG (0.5) and Blur+JPEG (0.1) variants, respectively. Furthermore, integrating our method with that of Gragnaniello et al.~\cite{Gragnaniello2021} yields a notable improvement of \(3.01\%\). These are testaments to the effectiveness of frequency masking, as it enables the model to learn more generalizable features for universal deepfake detection that are potentially overlooked by Wang et al.'s~\cite{wang2019cnngenerated} and Gragnaniello et al.'s~\cite{Gragnaniello2021} approaches.

Moreover, our approach also shines when coupled with 
Ojha et al.'s~\cite{ojha2023fakedetect} method which is based on visual encoder of CLIP~\cite{Radford2021LearningTV}, manifesting in a significant mAP boost of \(0.68\%\). This further affirms the adaptability of our approach across different backbones for universal deepfake detection. By strategically masking random frequencies, our model learns more generalizable features. Thus, these results shows our approach as valuable asset in the pursuit of more precise universal deepfake detection.

\vspace{-0.35cm}
\section{Conclusion}
\label{sec:conclusion}
\vspace{-0.2cm}

Motivated by recent promising results of masked image modeling, we proposed a frequency-based masking strategy tailored for universal detection of deepfakes.
Our method enables deepfake detector to learn generalizable features in the frequency domain.
Our experiments, conducted using samples with a range of generative models, empirically demonstrate the effectiveness of our proposed approach. 
The superior performance of our method 
highlights the potential of masked image modeling and frequency domain approach for  
the challenging problem of universal deepfake detection.

\vspace{-0.32cm}
\section{Acknowledgement}
\vspace{-0.2cm}

This research is supported by the National Research Foundation, Singapore under its AI Singapore Programmes (AISG Award No.: AISG2-TC-2022-007) and SUTD project PIE-SGP-AI-2018-01. This research work is also supported by the Agency for Science, Technology and Research (A*STAR) under its MTC Programmatic Funds (Grant No. M23L7b0021). This material is based on the research/work support in part by the Changi General Hospital and Singapore University of Technology and Design, under the HealthTech Innovation Fund (HTIF Award No. CGH-SUTD-2021-004). We thank anonymous reviewers for their insightful feedback.

\bibliographystyle{IEEEbib}
\bibliography{strings,refs}

\end{document}